# Modified CMA-ES Algorithm for Multi-Modal Optimization: Incorporating Niching Strategies and Dynamic Adaptation Mechanism

Wathsala Karunarathne[1], Indu Bala[2], Dikshit Chauhan[3], Matthew Roughan[1,2], Lewis Mitchell[2]

[1] Teletraffic Research Centre, The University of Adelaide, Australia
[2] The School of Computer and Mathematical Sciences, The University of Adelaide, Australia
[3] Department of Mathematics and Computing, National Institute of Technology, Punjab, India,

**Abstract:** This study modifies the Covariance Matrix Adaptation Evolution Strategy (CMA-ES) algorithm for multi-modal optimization problems. The enhancements focus on addressing the challenges of multiple global minima, improving the algorithm's ability to maintain diversity and explore complex fitness landscapes. We incorporate niching strategies and dynamic adaptation mechanisms to refine the algorithm's performance in identifying and optimizing multiple global optima. The algorithm generates a population of candidate solutions by sampling from a multivariate normal distribution centered around the current mean vector, with the spread determined by the step size and covariance matrix. Each solution's fitness is evaluated as a weighted sum of its contributions to all global minima, maintaining population diversity and preventing premature convergence. We implemented the algorithm on 8 tunable composite functions for the GECCO 2024 Competition on Benchmarking Niching Methods for Multi-Modal Optimization (MMO), adhering to the competition's benchmarking framework. The results are presenting in many ways such as Peak Ratio, F1 score on various dimensions. They demonstrate the algorithm's robustness and effectiveness in handling both global optimization and MMO-specific challenges, providing a comprehensive solution for complex multi-modal optimization problems.

**Introduction:** Many problems in industry, national defence, medicine, and energy can be transformed into real-parameter optimization problems [1-6]. These problems are universally acknowledged as important and challenging. Traditional gradient-based optimization algorithms can converge quickly in convex optimization but rely on differentiable and convex features of the problem. Additionally, these algorithms require a good initial guess for effective execution. However, most real-world engineering problems have become more complex, featuring multimodal, non- separable, nonconvex, and large-scale characteristics. As a result, deterministic algorithms struggle to address these problems effectively [7].

On the other hand, evolutionary algorithms (EAs) have proven effective at generating optimal or near-optimal solutions for such NP-hard problems [8]. EAs are pervasive across various scientific and engineering disciplines, often characterized by complex, high-dimensional, and non-linear landscapes. These algorithms mimic the principles of natural selection and genetics, making them robust and versatile methods for tackling such problems.

Among evolutionary algorithms, the derandomized Evolution Strategy (ES) with Covariance Matrix Adaptation (CMA) is renowned for its effectiveness in adapting the search distribution to the optimization landscape [9]. CMA iteratively refines a population of candidate solutions to approach the optimal solution. Initially designed to adapt the entire covariance matrix of the normal mutation distribution, CMA maintains several key invariances: order-preserving transformations of the objective function, angle-preserving transformations of the search space (e.g., rotation, reflection, translation), and scale invariance, provided initial conditions are appropriately set. These invariances ensure consistent behavior across various function classes, enhancing the generalizability of results. Originally intended for small population sizes, CMA-ES excels as a robust local search strategy, effectively minimizing unimodal test functions and handling ill-conditioned and non-separable problems.

Despite the success of CMA-ES, many real-world optimization problems are inherently multi-modal, featuring numerous local and global optima. Consequently, CMA-ES often struggles with these landscapes and tends to converge prematurely to suboptimal solutions [10]. Its tendency to converge towards a single optimum can lead to a loss of diversity, making it difficult to escape local optima and explore other promising regions of the search space [11-13]. To address this gap, this research proposes modifications to the traditional CMA-ES framework. By integrating niching strategies and dynamic adaptation mechanisms, the enhanced CMA-ES is designed to better handle multi-modal landscapes, maintain diversity, and effectively navigate complex search spaces.

This research aims to enhance the performance of CMA-ES in multi-modal optimization settings by introducing a modified version of the algorithm. The key contributions include:
- The incorporation of niching strategies to maintain population diversity and enable the exploration of multiple optima.
- The implementation of dynamic population size scaling to balance exploration and exploitation effectively.

These modifications collectively improve the algorithm's ability to identify and refine multiple optima in complex, multi-modal landscapes.

The remainder of this research is structured as follows: Section 2 provides a detailed explanation of the modified CMA-ES algorithm, including the mathematical formulations and implementation details. Section 3 presents the experimental setup and results, demonstrating the effectiveness of the proposed modifications. Finally, Section 4 concludes the paper with future research.

# SECTION 2 Method

This section presents a comprehensive description of the modified Covariance Matrix Adaptation Evolution Strategy (CMA-ES) algorithm tailored for multi-modal optimization problems. The modifications aim to address the challenges posed by multiple global minima, enhancing the algorithm's ability to maintain diversity and effectively explore complex fitness landscapes.

First the algorithm generates a population of candidate solutions by sampling from a multivariate normal distribution ($\mathcal{N}$) for the next generation ($t + 1$) centered around the current mean vector ($m$), with the spread determined by the step size ($\sigma$) and covariance matrix ($C$):

$$X_p^{t+1} \sim \mathcal{N}(X_w^t, \sigma^{t^2} C^t), p = 1, \ldots, \lambda \tag{1}$$

Such that

$$X_w^t = \sum_{j=1}^{\mu} w_j X_{j:\lambda}^t \tag{2}$$

where $\mathcal{N}(m, C)$ represents a random vector that follows a normal distribution with the mean $m$ and covariance matrix $C$. $X_w^t$ is the weighted average of the selected candidates, $w_j$ is positive weight as $\sum_{j=1}^{\mu} w_j = 1$. The notation $j: \lambda$ refers the $jth$ best candidates. We set all $w_j = \frac{1}{\mu}$, where $\mu$ is number of top-performing candidate solutions used to update the covariance matrix.

The evaluation of candidate solutions is customized to account for multiple global minima. Each solution's fitness is calculated as a weighted sum of its contributions to all global minima. This is achieved by applying the problem-specific shifts and rotations to each multimodal solution and evaluating it using a basic function tailored to the hardness of each global minimum. The weights for this summation are computed based on the proximity of the solution to each global minimum, normalized to ensure they sum to one.

Furthermore, the covariance matrix adaptation involves updating the evolution path for both the covariance matrix and the step size. The evolution path for the covariance matrix $P_c^{t+1}$ is computed using a combination of the previous path and the normalized step taken by the mean vector, adjusted by the covariance matrix:

$$P_c^{(t+1)} = (1 - c_c)P_c^t + h_\sigma^{(t+1)} \sqrt{c_c(2 - c_c)\mu_{eff}} \frac{X_w^{t+1} - X_w^t}{\sigma_t} \tag{3}$$

where the term $(1 - c_c)P_c^t$ reduces the influence of the previous evolution path by a factor $(1 - c_c)$. Where $c_c$ can be calculated as $c_c = \frac{4}{n+4}$. $n$ is the dimensionality of the optimization problem, representing the number of variables being optimized. $h_\sigma^{(t+1)}$ represents the indicator function is defined as follow:

$$h_\sigma^{(t+1)} = \begin{cases} 1 \text{ if } \frac{\|P_\sigma^{t+1}\|}{\sqrt{1-(1-c_\sigma)^{2(t+1)}}} < \left(1.5 + \frac{1}{n-0.5}\right) E(\|N(0,I)\|) \\ 0 \quad\quad\quad\quad\quad\quad\quad\quad\quad\quad\quad\quad\quad\quad\quad\quad otherwise \end{cases} \tag{4}$$

where $E(\|N(0,I)\|) = \sqrt{2 \frac{\Gamma\left(\frac{n+1}{2}\right)}{\Gamma\left(\frac{n}{2}\right)}}$ is expected length of $P_\sigma^{t+1}$ during random selection.

$$\mu_{eff} = \begin{cases} \mu \text{ if } w_j = \frac{1}{\mu} \\ \frac{1}{\sum_{j=1}^{\mu} w_j^2} \text{ otherwise} \end{cases} \quad (5)$$

The covariance matrix C itself is updated using $rank - one$ and $rank - \mu$ updates, which incorporate the evolution path and the weighted sum of the outer products of the selected parents' steps:

$$C^{t+1} = (1 - c_1 - c_\mu)C^t + c_1(P_c^{t+1}(P_c^{t+1})^T + \Delta C) + c_\mu \sum_{i=1}^{\mu} w_i \frac{X_{i:\lambda}^{t+1} - X_{i:\lambda}^t}{\sigma_t} \left(\frac{X_{i:\lambda}^{t+1} - X_{i:\lambda}^t}{\sigma_t}\right)^T \quad (6)$$

Where $c_1$ and $c_\mu$ are learning rate for $rank - one$ update and $rank - \mu$ update of the covariance matrix. $c_1$ controls the influence of the path and $c_\mu$ controls the influence of the selected candidate solutions on the covariance matrix on the covariance matrix. $\Delta$ is indicator function, which is 0 or 1 depending on whether step size control and path length conditions are met. This adaptation in equation (5) ensures that the covariance matrix accurately reflects the dependencies and correlations between variables, guiding the search towards promising regions.

Furthermore, the step size $\sigma$ is adapted using the cumulative step-size adaptation (CSA) mechanism. The evolution path for the step size $P_\sigma$ is updated based on the normalized steps taken by the mean vector. The step size is then adjusted exponentially, depending on the ratio of the norm of the evolution path to its expected value such as:

$$\sigma^{t+1} = \sigma^t \cdot \exp\left(\frac{c_\sigma}{d_\sigma}\left(\frac{\|P_\sigma^{t+1}\|}{E(\|N(0,I)\|)} - 1\right)\right) \quad (7)$$

Where $c_\sigma = \frac{2 + \mu_{eff}}{3 + n + \mu_{eff}}, d_\sigma = 1 + 2 \max\left(0, \sqrt{\frac{\mu_{eff} - 1}{n+1}} - 1\right) + c_\sigma$ \quad (8)

This adaptation balances exploration and exploitation, ensuring that the algorithm can both explore new regions of the search space and refine known good regions. The detailed pseudo code of implanting this strategy on multimodal problems are presented in Algorithm 1.

**Algorithm 1:** Modified Covariance Matrix Adaptation Evolution Strategy (CMA-ES) algorithm

> **Step 1:** Initialize problem-specific parameters for multimodal benchmark:
> Generate positions $(X_i)$ using uniform random numbers (consider as global solutions)
> Adjust positions for non-uniformity and map to search space bounds
> Calculate hardness for each point as $H_{glob,i} = \left(\frac{index_i - 1}{n(global \text{ minima}) - 1}\right) \times (\max\_hardness - \min hardness) + \min hardness$
> $for$ multimodal problems, compute niching radius $\sigma_{nich} = \min(|X_i, X_j|)/2$
> Generate rotation matrices for global and local minima

```
Step 2: Initialize CMA-ES parameters:
    Set initial mean vector $m_0$
    Set initial covariance matrix $C_0$ to identity matrix
    Set initial step size $\sigma_0$
    Determine population size $n$
    Calculate weights $w_i$ and number of parents $\mu$

while not converged:
    Generate lambda candidate solutions $X_p$ from multivariate normal distribution using equation (1)
    Apply shifts and rotations to each candidate solution
        for each candidate solution $X_p$:
        Calculate weights $w_p$ based on proximity to global minima:
          $w_p = exp\left(\frac{\sigma_X}{\sigma_w \times \sigma_{nich}}\right)^2 - C_0;\ C_0 = max\left(0, 1 - min\left(\frac{\sigma_X}{\sigma_w \times \sigma_{nich}}\right)^2\right)$
            Normalize weights:
        Calculate overall fitness:
Step 3: Select top $\mu$ candidates based on fitness values
        for each $i$ in range ($\mu$)
            Update mean vector $m$ using equation (2)
    Update covariance matrix ($C$):
        Calculate evolution path for covariance matrix $P_c^{t+1}$ using equation (3):
            Update the $rank - one$ and $rank - \mu$ using equations (4-6)
    Adapt step size ($\sigma_p^{t+1}$)
        Calculate evolution path for step size ($P_\sigma^{t+1}$)
        Update step size using equations (7-8)
    Update previous mean vector to current mean vector
Check for convergence criteria:
    if convergence criteria met:
        break loop
Return final optimized solutions
```

## SECTION 3: Experimental Results and Discussion

**3.1 Test Function:** We implemented the modified CMA-ES algorithm for the GECCO 2024 Competition on Benchmarking Niching Methods for Multimodal Optimization (MMO), adhering to the same settings defined in this competition's benchmarking framework. Our study utilized a set of 8 tuneable composite functions, specifically designed to overcome the limitations of the widely used CEC 2013 test problems for MMO [14], which include limited scalability, low dimensionality, and insufficient differentiation among competitive MMO methods. These functions were developed by adapting and enhancing procedures from previous studies and by categorizing MMO challenges into global optimization challenges and MMO-specific challenges.

Each of the eight composite functions was tested under two different parameter settings, resulting in 16 MMO problems classified into two groups:

- **Group A**: These problems evaluate an MMO method's ability to handle challenges specific to MMO, such as a higher number of global minima with non-uniform distribution.

- **Group B**: These problems assess an MMO method's ability to address both global optimization challenges (e.g., deceptive local minima, ill-conditioning, and weak global structure) and MMO-specific challenges.

The best fitness values and the number of global optima for these functions are presented in Table 1.

**Table 1:** GECCO'2024 Competition on Benchmarking Niching Methods for Multimodal Optimization Test functions

| Groups | Problem ID | Functions | Best Fitness value ($f^*$) | Number of Global Minima |
|---|---|---|---|---|
| Group A | 1 | High-Conditioned Elliptic | -97.8 | 20 |
| | 2 | Different Powers | 64.1 | 20 |
| | 3 | Skewed Schwefel No2 | 483.1 | 20 |
| | 4 | Rosenbrock | -96.7 | 20 |
| | 5 | Skewed Ackley | -395 | 20 |
| | 6 | Rastrigin | -34.6 | 20 |
| | 7 | Penalized Weierstrass | 494 | 20 |
| | 8 | Penalized Schwefel N26 | -402.2 | 20 |
| Group B | 9 | High-Conditioned Elliptic | -97.8 | 10 |
| | 10 | Different Powers | 64.1 | 10 |
| | 11 | Skewed Schwefel No2 | 483.1 | 10 |
| | 12 | Rosenbrock | -96.7 | 10 |
| | 13 | Skewed Ackley | -395 | 10 |
| | 14 | Rastrigin | -34.6 | 10 |
| | 15 | Penalized Weierstrass | 494 | 10 |
| | 16 | Penalized Schwefel N26 | -402.2 | 10 |

**3.2 Parameter Setting:** In this section, we discuss how we set the parameters for the CMA-ES algorithm and the additions we made to improve its accuracy and speed as a niching method for MMO. We apply this method across problem dimensionalities 2, 5, 10, and 20, with search space bounds set to [-5, 5] for each dimension. The main stopping criterion is a limited evaluation budget, where the maximum budget is determined by multiplying the problem dimension by 50,000.

A key enhancement we implemented to improve the accuracy and speed is initialising each iteration's search with the best solution from the previous iteration. We set the population size proportionally to the number of variables, and the number of parents is half the population size. Additionally, parameters for step size control and covariance update are tailored based on the number of variables. Using the number of variables in parameter settings offers several benefits:

- The performance of optimisation algorithms often depends on how well their parameters scale with the dimensionality of the problem. As the number of dimensions increases, some parameters may need adjustments to maintain efficiency and effectiveness. On the other hand, if the parameters are not properly set for the problem dimension, it can lead to computational cost, longer convergence times, or even algorithm failure [15].
- The nature of the search space in higher-dimensional problems differs significantly from lower-dimensional ones. Parameters such as step size control parameters, covariance update parameters, and population size parameters in CMA-ES algorithm need to be adjusted to adequately explore the larger space without being too conservative or too aggressive.
- Parameters that are tailored to the problem dimension can help algorithms converge faster and find higher-quality solutions. Further, properly set parameters can mitigate the risk of premature convergence by encountering sufficient exploration of the search space.
- Parameters that are well-suited to the problem dimension contribute to the algorithm's robustness as well.

**3.3 Experimental Results:** We generated results for 16 test problems (8 from Group A and 8 from Group B) presented in Table 1, across 15 instances for each of the 4 different dimension values (2, 5, 10, 20). The total number of optimisation runs was $16 \times 15 \times 4 = 960$.

We use two performance matrices to illustrate the capabilities of the CMA-ES algorithm as a niching method for MMO. The first one is the well-known peak ratio (PR), defined as $\epsilon_f$, which presents the accuracy difference between the actual function value ($f^*$) and the detected function value ($f(x)$) from the optimisation run [12].

$$\epsilon_f = f(x) - f^* \tag{9}$$

We recorded $\epsilon_f$ for the 15 different problem instances. The problems listed in Table 1 are deterministic; however, the procedure involves some randomness. Therefore, these 15 instances are used to mitigate the impact of this randomness. Table 1 presents the results for 16 GECCO benchmark functions for problem instance 1, with detailed results for the other instances provided in the Appendix.

**Table 1:** Results of multimodal functions for dimensions 2, 5, 10, and 20 for problem instance 1.

| Problem ID | $f^*$ | Dimension =2 | | Dimension=5 | | Dimension=10 | | Dimension=20 | |
|---|---|---|---|---|---|---|---|---|---|
| | | $f(x)$ | $\epsilon_f$ | $f(x)$ | $\epsilon_f$ | $f(x)$ | $\epsilon_f$ | $f(x)$ | $\epsilon_f$ |
| 1 | −97.8 | −97.7999 | 0.0000 | −97.7999 | 0.0000 | −97.7999 | 0.0000 | −97.7999 | 0.0000 |
| 2 | 64.1 | 64.1000 | 0.0000 | 64.1000 | 0.0000 | 64.1000 | 0.0000 | 64.1002 | 0.0002 |
| 3 | 483.1 | 483.1000 | 0.0000 | 483.1000 | 0.0000 | 483.1001 | 0.0001 | 497.5441 | 14.4441 |
| 4 | −96.7 | −96.6999 | 0.0000 | −96.6999 | 0.0000 | −96.6999 | 0.0000 | −96.6999 | 0.0000 |
| 5 | −395 | −394.9999 | 0.0000 | −394.9999 | 0.0000 | −394.9999 | 0.00008 | −394.9993 | 0.0006 |
| 6 | −34.6 | −34.6 | 0.0000 | −34.5999 | 0.0000 | −34.6 | 0.0000 | −34.6 | 0.0000 |
| 7 | 494 | 494.0000 | 0.0000 | 494.0000 | 0.0000 | 494.0000 | 0.0000 | 494.0000 | 0.0001 |
| 8 | −402.2 | −402.1999 | 0.0000 | −402.1998 | 0.0001 | −402.1929 | 0.0071 | −402.07716 | 0.1228 |
| 9 | −97.8 | −97.7999 | 0.0000 | −97.7999 | 0.0000 | −97.7999 | 0.0000 | −97.7999 | 0.0000 |
| 10 | 64.1 | 64.1000 | 0.0000 | 64.1000 | 0.0000 | 64.1000 | 0.0000 | 64.1002 | 0.0002 |
| 11 | 483.1 | 483.1000 | 0.0000 | 483.1000 | 0.0000 | 483.1000 | 0.0000 | 495.02224 | 11.9222 |
| 12 | −96.7 | −96.7 | 0.0000 | −96.6999 | 0.0000 | −96.6999 | 0.0000 | −96.6999 | 0.0000 |
| 13 | −395 | −394.9999 | 0.0000 | −394.9999 | 0.0000 | −394.9999 | 0.00007 | −394.9993 | 0.0006 |
| 14 | −34.6 | −34.6 | 0.0000 | −34.5999 | 0.0001 | −34.5999 | 0.0001 | −34.5999 | 0.0001 |
| 15 | 494.0000 | 494.0000 | 0.0000 | 494.0000 | 0.0000 | 494.0000 | 0.0000 | 494.0001 | 0.0001 |

| 16 | −402.2 | −402.1999 | 0.00009 | −402.1998 | 0.001 | −402.1919 | 0.008 | −402.0525 | 0.1474 |

The second performance metric is the $F_1$ score, defined as:

$$F_1 = 2 \times \frac{Precision \times Recall}{Precision + Recall} \tag{10}$$

Here, recall in the context of MMO represents the fraction of global minima that have been detected by the algorithm among the existing global minima, equivalent to $\epsilon_f$. Precision is the ratio of the number of detected global minima by the algorithm to the number of reported global minima. The average precision is 0.869, and the average $F_1$ score is 0.6682 for the 960 optimization runs. The overall score, calculated based on $\epsilon_f$ and the $F_1$ score, is 0.6055, demonstrating the modified CMA-ES algorithm's high performance as a niching method.

Figures 1 illustrate the algorithm's convergence for four different functions (IDs 9, 11, 14, and 16). The algorithm converges quickly to the provided optimal values $f^*$, with peak ratios $\epsilon_f$ less than 0.01 in these cases. In these plots function value decreases rapidly within the first few iterations, indicating quick convergence. Furthermore, after the initial drop, the function value stabilizes, suggesting that the algorithm has reached or is very close to the optimal value. This indicates that the modified CMA-ES algorithm is effective in quickly finding the optimal solution.

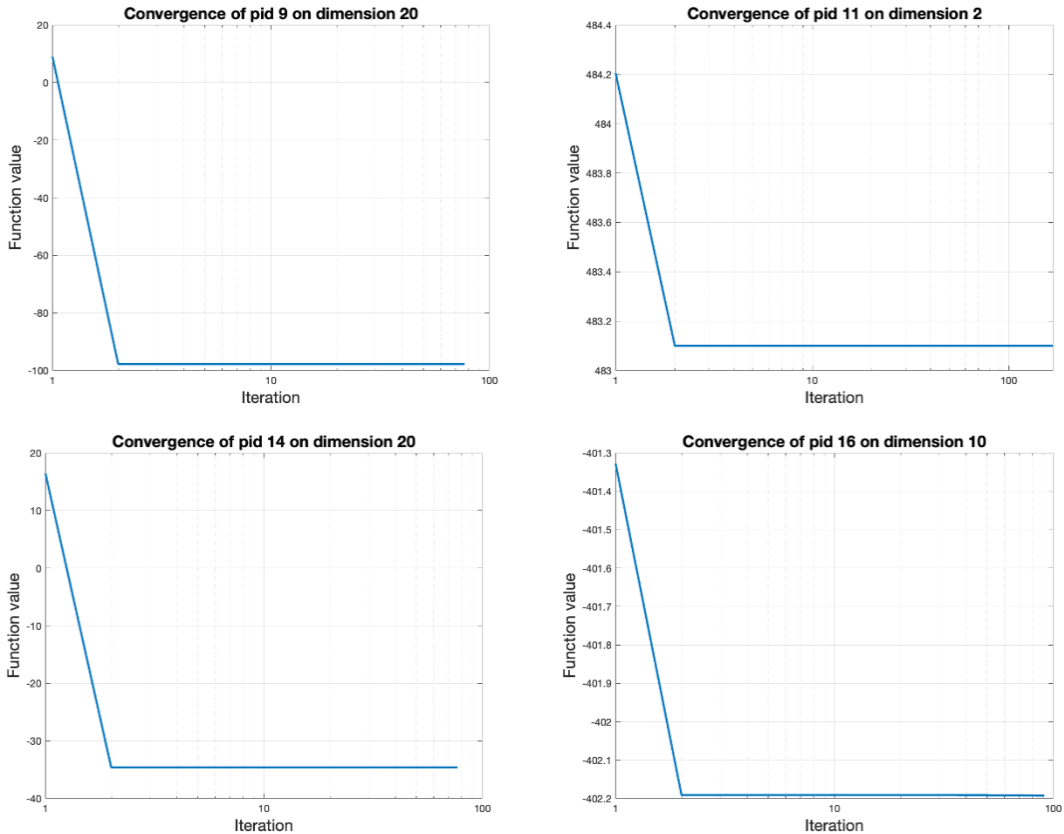

**Figure 1:** Convergence plots of the modified CMA-ES algorithm for function IDs 9, 11, 14, and 16 across various dimensions.

**SECTION 4:** Conclusion and Future direction

In this study, we implemented a modified Covariance Matrix Adaptation Evolution Strategy (CMA-ES) tailored for GECCO multi-modal optimization problems. By incorporating niching strategies and dynamic adaptation mechanisms, our enhanced algorithm effectively maintains diversity and navigates complex search landscapes. We evaluated the modified CMA-ES on 16 GECCO benchmark functions across multiple dimensions and problem instances. The results demonstrate that the modified CMA-ES can find all global optima in challenging multi-modal scenarios. The integration of niching strategies helps preserve population diversity, preventing premature convergence to suboptimal solutions. The dynamic adaptation mechanisms ensure that the algorithm can balance exploration and exploitation, adapting the search distribution to the underlying problem landscape.

While the modified CMA-ES has shown promising results, our future intention is to apply this algorithm to other benchmarks and compare it with existing algorithms in terms of convergence complexity and computational time. Applying the modified CMA-ES to a broader range of real-world problems in domains such as engineering, medicine, and finance will help validate its practical utility and robustness. Additionally, implementing the modified CMA-ES in parallel and distributed computing environments could significantly reduce computation time and enable the handling of larger, more complex problems. By addressing these future directions, we can further enhance the capabilities of CMA-ES and expand its applicability to a wider range of optimization challenges.

# Appendix

**A1:** Results of multimodal functions for dimensions 2, 5, 10, and 20 for problem instance 2.

| Problem ID | $f^*$ | Dimension =2 | | Dimension=5 | | Dimension=10 | | Dimension=20 | |
|---|---|---|---|---|---|---|---|---|---|
| | | $f(x)$ | $\epsilon_f$ | $f(x)$ | $\epsilon_f$ | $f(x)$ | $\epsilon_f$ | $f(x)$ | $\epsilon_f$ |
| 1 | −97.8 | −97.8000 | 0.0000 | −97.7999 | 0.0000 | −97.7999 | 0.0000 | −97.7999 | 0.0000 |
| 2 | 64.1 | 64.1000 | 0.0000 | 64.1000 | 0.0000 | 64.1000 | 0.0001 | 64.1002 | 0.0002 |
| 3 | 483.1 | 483.1000 | 0.0000 | 483.1000 | 0.0000 | 483.1005 | 0.0005 | 505.9763 | 22.8763 |
| 4 | −96.7 | −96.6999 | 0.0000 | −96.6999 | 0.0000 | −96.6999 | 0.0000 | −96.6999 | 0.0000 |
| 5 | −395 | −394.9999 | 0.0000 | −394.9999 | 0.0000 | −394.9999 | 0.00006 | −394.9994 | 0.0005 |
| 6 | −34.6 | −34.5999 | 0.0000 | −34.5999 | 0.0000 | −34.5999 | 0.0000 | −34.5999 | 0.0000 |
| 7 | 494 | 494.0000 | 0.0000 | 494.0000 | 0.0000 | 494.0000 | 0.0000 | 494.0001 | 0.0001 |
| 8 | −402.2 | −402.1999 | 0.00001 | −402.1998 | 0.0001 | −402.1919 | 0.0080 | −402.0561 | 0.1438 |
| 9 | −97.8 | −97.7999 | 0.0000 | −97.7999 | 0.0000 | −97.7999 | 0.0000 | −97.7999 | 0.0000 |
| 10 | 64.1 | 64.1000 | 0.0000 | 64.1000 | 0.0000 | 64.1000 | 0.0000 | 64.1002 | 0.0002 |
| 11 | 483.1 | 483.1000 | 0.0000 | 483.1000 | 0.0000 | 483.1001 | 0.0001 | 494.5694 | 11.4694 |
| 12 | −96.7 | −96.7000 | 0.0000 | −96.6999 | 0.0000 | −96.6999 | 0.0000 | −96.6999 | 0.0000 |
| 13 | −395 | −394.9999 | 0.0000 | −394.9999 | 0.0000 | −394.9998 | 0.0001 | −394.9992 | 0.0007 |
| 14 | −34.6 | −34.5999 | 0.0000 | −34.5999 | 0.0000 | −34.5999 | 0.0000 | −34.5999 | 0.0000 |
| 15 | 494 | 494.0000 | 0.0000 | 494.0000 | 0.0000 | 494.0000 | 0.0000 | 494.0001 | 0.0001 |
| 16 | −402.2 | −402.1999 | 0.00005 | −402.1924 | 0.0075 | −402.0561 | 0.1438 | −402.0525 | 0.1474 |

**A2:** Results of multimodal functions for dimensions 2, 5, 10, and 20 for problem instance 3.

| Problem ID | $f^*$ | Dimension =2 | | Dimension=5 | | Dimension=10 | | Dimension=20 | |
|---|---|---|---|---|---|---|---|---|---|
| | | $f(x)$ | $\epsilon_f$ | $f(x)$ | $\epsilon_f$ | $f(x)$ | $\epsilon_f$ | $f(x)$ | $\epsilon_f$ |
| 1 | −97.8 | −97.8000 | 0.0000 | −97.7999 | 0.0000 | −97.7999 | 0.0000 | −97.7999 | 0.0000 |
| 2 | 64.1 | 64.1000 | 0.0000 | 64.1000 | 0.0000 | 64.1000 | 0.0001 | 64.1002 | 0.0002 |
| 3 | 483.1 | 483.1000 | 0.0000 | 483.1000 | 0.0000 | 483.1003 | 0.0003 | 501.5398 | 18.4398 |
| 4 | −96.7 | −96.6999 | 0.0000 | −96.6999 | 0.0000 | −96.6999 | 0.0000 | −96.6999 | 0.0000 |
| 5 | −395 | −394.9999 | 0.0000 | −394.9999 | 0.0000 | −394.9999 | 0.00007 | −394.9993 | 0.0006 |
| 6 | −34.6 | −34.5999 | 0.0000 | −34.5999 | 0.0000 | −34.5999 | 0.0000 | −34.5999 | 0.0000 |
| 7 | 494 | 494.0000 | 0.0000 | 494.0000 | 0.0000 | 494.0000 | 0.0000 | 494.0001 | 0.0001 |
| 8 | −402.2 | −402.1999 | 0.00001 | −402.1998 | 0.0001 | −402.1917 | 0.0082 | −402.0680 | 0.1319 |
| 9 | −97.8 | −97.7999 | 0.0000 | −97.7999 | 0.0000 | −97.7999 | 0.0000 | −97.7999 | 0.0000 |
| 10 | 64.1 | 64.1000 | 0.0000 | 64.1000 | 0.0000 | 64.1000 | 0.00001 | 64.1002 | 0.0002 |
| 11 | 483.1 | 483.1000 | 0.0000 | 483.1000 | 0.0000 | 483.1000 | 0.00005 | 499.2911 | 16.1911 |
| 12 | −96.7 | −96.6999 | 0.0000 | −96.6999 | 0.0000 | −96.6999 | 0.0000 | −96.6999 | 0.0000 |
| 13 | −395 | −394.9999 | 0.0000 | −394.9999 | 0.0000 | −394.9998 | 0.0001 | −394.9993 | 0.0006 |
| 14 | −34.6 | −34.5999 | 0.0000 | −34.5999 | 0.0000 | −34.5999 | 0.0000 | −34.5999 | 0.0000 |
| 15 | 494 | 494.0000 | 0.0000 | 494.0000 | 0.0000 | 494.0000 | 0.0000 | 494.0001 | 0.0001 |
| 16 | −402.2 | −402.1999 | 0.00005 | −402.1998 | 0.0001 | −402.1940 | 0.0059 | −402.0837 | 0.1162 |

**A3:** Results of multimodal functions for dimensions 2, 5, 10, and 20 for problem instance 4.

| Problem ID | $f^*$ | Dimension =2 | | Dimension=5 | | Dimension=10 | | Dimension=20 | |
|---|---|---|---|---|---|---|---|---|---|
| | | $f(x)$ | $\epsilon_f$ | $f(x)$ | $\epsilon_f$ | $f(x)$ | $\epsilon_f$ | $f(x)$ | $\epsilon_f$ |
| 1 | −97.8 | −97.7999 | 0.0000 | −97.7999 | 0.0000 | −97.7999 | 0.0000 | −97.7999 | 0.0000 |
| 2 | 64.1 | 64.1000 | 0.0000 | 64.1000 | 0.0000 | 64.1000 | 0.0000 | 64.1002 | 0.0002 |
| 3 | 483.1 | 483.1000 | 0.0000 | 483.1000 | 0.0000 | 483.1010 | 0.0010 | 492.9733 | 9.8733 |
| 4 | −96.7 | −96.6999 | 0.0000 | −96.6999 | 0.0000 | −96.6999 | 0.0000 | −96.6999 | 0.0000 |
| 5 | −395 | −394.9999 | 0.0000 | −394.9999 | 0.0000 | −394.9998 | 0.0001 | −394.9993 | 0.0006 |
| 6 | −34.6 | −34.5999 | 0.0000 | −34.5999 | 0.0000 | −34.5999 | 0.0000 | −34.5999 | 0.0000 |
| 7 | 494 | 494.0000 | 0.0000 | 494.0000 | 0.0000 | 494.0000 | 0.0000 | 494.0001 | 0.0001 |
| 8 | −402.2 | −402.1999 | 0.0000 | −402.1999 | 0.00009 | −402.1928 | 0.0071 | −402.0739 | 0.1260 |
| 9 | −97.8 | −97.7999 | 0.0000 | −97.7999 | 0.0000 | −97.7999 | 0.0000 | −97.7999 | 0.0000 |
| 10 | 64.1 | 64.1000 | 0.0000 | 64.1000 | 0.0000 | 64.1000 | 0.00001 | 64.1003 | 0.0003 |
| 11 | 483.1 | 483.1000 | 0.0000 | 483.1000 | 0.0000 | 483.1003 | 0.0003 | 500.5302 | 17.4302 |
| 12 | −96.7 | −96.6999 | 0.0000 | −96.6999 | 0.0000 | −96.6999 | 0.0000 | −96.6999 | 0.0000 |
| 13 | −395 | −394.9999 | 0.0000 | −394.9999 | 0.0000 | −394.9999 | 0.00005 | −394.9992 | 0.0007 |
| 14 | −34.6 | −34.5999 | 0.0000 | −34.5999 | 0.0000 | −34.5999 | 0.0000 | −34.5999 | 0.0000 |
| 15 | 494 | 494.0000 | 0.0000 | 494.0000 | 0.0000 | 494.0000 | 0.0000 | 494.0001 | 0.0001 |
| 16 | −402.2 | −402.1999 | 0.00005 | −402.1998 | 0.0002 | −402.1917 | 0.0082 | −402.0835 | 0.1164 |

**A4:** Results of multimodal functions for dimensions 2, 5, 10, and 20 for problem instance 5.

| Problem ID | $f^*$ | Dimension =2 | | Dimension=5 | | Dimension=10 | | Dimension=20 | |
|---|---|---|---|---|---|---|---|---|---|
| | | $f(x)$ | $\epsilon_f$ | $f(x)$ | $\epsilon_f$ | $f(x)$ | $\epsilon_f$ | $f(x)$ | $\epsilon_f$ |
| 1 | −97.8 | −97.7999 | 0.0000 | −97.7999 | 0.0000 | −97.7999 | 0.0000 | −97.7999 | 0.0000 |
| 2 | 64.1 | 64.1000 | 0.0000 | 64.1000 | 0.0000 | 64.1000 | 0.00001 | 64.1002 | 0.0002 |
| 3 | 483.1 | 483.1000 | 0.0000 | 483.1000 | 0.0000 | 483.1006 | 0.0006 | 494.7228 | 11.6228 |
| 4 | −96.7 | −96.6999 | 0.0000 | −96.6999 | 0.0000 | −96.6999 | 0.0000 | −96.6999 | 0.0000 |
| 5 | −395 | −394.9999 | 0.0000 | −394.9999 | 0.0000 | −394.9999 | 0.00006 | −394.9993 | 0.0006 |
| 6 | −34.6 | −34.5999 | 0.0000 | −34.5999 | 0.0000 | −34.5999 | 0.0000 | −34.5999 | 0.0000 |
| 7 | 494 | 494.0000 | 0.0000 | 494.0000 | 0.0000 | 494.0000 | 0.0000 | 494.0001 | 0.0001 |
| 8 | −402.2 | −402.1999 | 0.0000 | −402.1997 | 0.0002 | −402.1941 | 0.0058 | −402.0569 | 0.1430 |
| 9 | −97.8 | −97.7999 | 0.0000 | −97.7999 | 0.0000 | −97.7999 | 0.0000 | −97.7999 | 0.0000 |
| 10 | 64.1 | 64.1000 | 0.0000 | 64.1000 | 0.0000 | 64.1000 | 0.00002 | 64.1002 | 0.0002 |
| 11 | 483.1 | 483.1000 | 0.0000 | 483.1000 | 0.0000 | 483.1000 | 0.00003 | 507.3025 | 24.2025 |
| 12 | −96.7 | −96.6999 | 0.0000 | −96.6999 | 0.0000 | −96.6999 | 0.0000 | −96.6999 | 0.0000 |
| 13 | −395 | −394.9999 | 0.0000 | −394.9999 | 0.0000 | −394.9999 | 0.00008 | −394.9993 | 0.0006 |
| 14 | −34.6 | −34.5999 | 0.0000 | −34.5999 | 0.0000 | −34.5999 | 0.0000 | −34.5999 | 0.0000 |
| 15 | 494 | 494.0000 | 0.0000 | 494.0000 | 0.0000 | 494.0000 | 0.0000 | 494.0001 | 0.0001 |
| 16 | −402.2 | −402.1999 | 0.0000 | −402.1998 | 0.0002 | −402.1917 | 0.0082 | −402.0523 | 0.1476 |

**A5:** Results of multimodal functions for dimensions 2, 5, 10, and 20 for problem instance 6.

| Problem ID | $f^*$ | Dimension =2 | | Dimension=5 | | Dimension=10 | | Dimension=20 | |
|---|---|---|---|---|---|---|---|---|---|
| | | $f(x)$ | $\epsilon_f$ | $f(x)$ | $\epsilon_f$ | $f(x)$ | $\epsilon_f$ | $f(x)$ | $\epsilon_f$ |
| 1 | −97.8 | −97.7999 | 0.0000 | −97.7999 | 0.0000 | −97.7999 | 0.0000 | −97.7999 | 0.0000 |
| 2 | 64.1 | 64.1000 | 0.0000 | 64.1000 | 0.0000 | 64.1000 | 0.00001 | 64.1002 | 0.0002 |
| 3 | 483.1 | 483.1000 | 0.0000 | 483.1000 | 0.0000 | 483.1004 | 0.0004 | 507.0331 | 23.9331 |
| 4 | −96.7 | −96.6999 | 0.0000 | −96.6999 | 0.0000 | −96.6999 | 0.0000 | −96.6999 | 0.0000 |
| 5 | −395 | −394.9999 | 0.0000 | −394.9999 | 0.0000 | −394.9999 | 0.00005 | −394.9994 | 0.0005 |
| 6 | −34.6 | −34.5999 | 0.0000 | −34.5999 | 0.0000 | −34.5999 | 0.0000 | −34.5999 | 0.0000 |
| 7 | 494 | 494.0000 | 0.0000 | 494.0000 | 0.0000 | 494.0000 | 0.0000 | 494.0001 | 0.0001 |
| 8 | −402.2 | −402.1999 | 0.0000 | −402.1998 | 0.0001 | −402.1948 | 0.0051 | −402.0491 | 0.1508 |
| 9 | −97.8 | −97.7999 | 0.0000 | −97.7999 | 0.0000 | −97.7999 | 0.0000 | −97.7999 | 0.0000 |
| 10 | 64.1 | 64.1000 | 0.0000 | 64.1000 | 0.0000 | 64.1000 | 0.00001 | 64.1002 | 0.0002 |
| 11 | 483.1 | 483.1000 | 0.0000 | 483.1000 | 0.0000 | 483.1000 | 0.0000 | 493.4287 | 10.3287 |
| 12 | −96.7 | −96.6999 | 0.0000 | −96.6999 | 0.0000 | −96.6999 | 0.0000 | −96.6999 | 0.0000 |
| 13 | −395 | −394.9999 | 0.0000 | −394.9999 | 0.0000 | −394.9999 | 0.00007 | −394.9994 | 0.0005 |
| 14 | −34.6 | −34.5999 | 0.0000 | −34.5999 | 0.0000 | −34.5999 | 0.0000 | −34.5999 | 0.0000 |
| 15 | 494 | 494.0000 | 0.0000 | 494.0000 | 0.0000 | 494.0000 | 0.0000 | 494.0001 | 0.0001 |
| 16 | −402.2 | −402.1999 | 0.00003 | −402.1998 | 0.0001 | −402.1947 | 0.0052 | −402.0541 | 0.1458 |

**A6:** Results of multimodal functions for dimensions 2, 5, 10, and 20 for problem instance 7.

| Problem ID | $f^*$ | Dimension =2 | | Dimension=5 | | Dimension=10 | | Dimension=20 | |
|---|---|---|---|---|---|---|---|---|---|
| | | $f(x)$ | $\epsilon_f$ | $f(x)$ | $\epsilon_f$ | $f(x)$ | $\epsilon_f$ | $f(x)$ | $\epsilon_f$ |
| 1 | −97.8 | −97.7999 | 0.0000 | −97.7999 | 0.0000 | −97.7999 | 0.0000 | −97.7999 | 0.0000 |
| 2 | 64.1 | 64.1000 | 0.0000 | 64.1000 | 0.0000 | 64.1000 | 0.00001 | 64.1002 | 0.0002 |
| 3 | 483.1 | 483.1000 | 0.0000 | 483.1000 | 0.0000 | 483.1000 | 0.00006 | 494.0010 | 10.9010 |
| 4 | −96.7 | −96.6999 | 0.0000 | −96.6999 | 0.0000 | −96.6999 | 0.0000 | −96.6999 | 0.0000 |
| 5 | −395 | −394.9999 | 0.0000 | −394.9999 | 0.0000 | −394.9999 | 0.00006 | −394.9992 | 0.0007 |
| 6 | −34.6 | −34.5999 | 0.0000 | −34.5999 | 0.0000 | −34.5999 | 0.0000 | −34.5999 | 0.0000 |
| 7 | 494 | 494.0000 | 0.0000 | 494.0000 | 0.0000 | 494.0000 | 0.0000 | 494.0001 | 0.0001 |
| 8 | −402.2 | −402.1999 | 0.0000 | −402.1998 | 0.0001 | −402.1919 | 0.0080 | −402.0682 | 0.1317 |
| 9 | −97.8 | −97.7999 | 0.0000 | −97.7999 | 0.0000 | −97.7999 | 0.0000 | −97.7999 | 0.0000 |
| 10 | 64.1 | 64.1000 | 0.0000 | 64.1000 | 0.0000 | 64.1000 | 0.00001 | 64.1001 | 0.0001 |
| 11 | 483.1 | 483.1000 | 0.0000 | 483.1000 | 0.0000 | 483.1000 | 0.0000 | 499.0443 | 15.9443 |
| 12 | −96.7 | −96.6999 | 0.0000 | −96.6999 | 0.0000 | −96.6999 | 0.0000 | −96.6999 | 0.0000 |
| 13 | −395 | −394.9999 | 0.0000 | −394.9999 | 0.0000 | −394.9999 | 0.00007 | −394.9992 | 0.0007 |
| 14 | −34.6 | −34.5999 | 0.0000 | −34.5999 | 0.0000 | −34.5999 | 0.0000 | −34.5999 | 0.0000 |
| 15 | 494 | 494.0000 | 0.0000 | 494.0000 | 0.0000 | 494.0000 | 0.0000 | 494.0001 | 0.0001 |
| 16 | −402.2 | −402.1999 | 0.0000 | −402.1998 | 0.0001 | −402.1916 | 0.0083 | −402.0581 | 0.1481 |

**A7:** Results of multimodal functions for dimensions 2, 5, 10, and 20 for problem instance 8.

| Problem ID | $f^*$ | Dimension =2 | | Dimension=5 | | Dimension=10 | | Dimension=20 | |
|---|---|---|---|---|---|---|---|---|---|
| | | $f(x)$ | $\epsilon_f$ | $f(x)$ | $\epsilon_f$ | $f(x)$ | $\epsilon_f$ | $f(x)$ | $\epsilon_f$ |
| 1 | −97.8 | −97.7999 | 0.0000 | −97.7999 | 0.0000 | −97.7999 | 0.0000 | −97.7999 | 0.0000 |
| 2 | 64.1 | 64.1000 | 0.0000 | 64.1000 | 0.0000 | 64.1000 | 0.00001 | 64.1002 | 0.0002 |
| 3 | 483.1 | 483.1000 | 0.0000 | 483.1000 | 0.0000 | 483.1000 | 0.00005 | 507.2542 | 24.1542 |
| 4 | −96.7 | −96.6999 | 0.0000 | −96.6999 | 0.0000 | −96.6999 | 0.0000 | −96.6999 | 0.0000 |
| 5 | −395 | −394.9999 | 0.0000 | −394.9999 | 0.0000 | −394.9999 | 0.00007 | −394.9993 | 0.0006 |
| 6 | −34.6 | −34.5999 | 0.0000 | −34.5999 | 0.0000 | −34.5999 | 0.0000 | −34.5999 | 0.0000 |
| 7 | 494 | 494.0000 | 0.0000 | 494.0000 | 0.0000 | 494.0000 | 0.0000 | 494.0001 | 0.0001 |
| 8 | −402.2 | −402.1999 | 0.0000 | −402.1998 | 0.0001 | −402.1909 | 0.0090 | −402.0574 | 0.1425 |
| 9 | −97.8 | −97.7999 | 0.0000 | −97.7999 | 0.0000 | −97.7999 | 0.0000 | −97.7999 | 0.0000 |
| 10 | 64.1 | 64.1000 | 0.0000 | 64.1000 | 0.0000 | 64.1000 | 0.00001 | 64.1002 | 0.0002 |
| 11 | 483.1 | 483.1000 | 0.0000 | 483.1000 | 0.0000 | 483.1012 | 0.012 | 509.1816 | 26.0816 |
| 12 | −96.7 | −96.6999 | 0.0000 | −96.6999 | 0.0000 | −96.6999 | 0.0000 | −96.6999 | 0.0000 |
| 13 | −395 | −394.9999 | 0.0000 | −394.9999 | 0.0000 | −394.9999 | 0.00007 | −394.9993 | 0.0006 |
| 14 | −34.6 | −34.5999 | 0.0000 | −34.5999 | 0.0000 | −34.5999 | 0.0000 | −34.5999 | 0.0000 |
| 15 | 494 | 494.0000 | 0.0000 | 494.0000 | 0.0000 | 494.0000 | 0.0000 | 494.0001 | 0.0001 |
| 16 | −402.2 | −402.1999 | 0.0000 | −402.1998 | 0.0001 | −402.1929 | 0.0070 | −402.1048 | 0.0951 |

**A8:** Results of multimodal functions for dimensions 2, 5, 10, and 20 for problem instance 9.

| Problem ID | $f^*$ | Dimension =2 | | Dimension=5 | | Dimension=10 | | Dimension=20 | |
|---|---|---|---|---|---|---|---|---|---|
| | | $f(x)$ | $\epsilon_f$ | $f(x)$ | $\epsilon_f$ | $f(x)$ | $\epsilon_f$ | $f(x)$ | $\epsilon_f$ |
| 1 | −97.8 | −97.7999 | 0.0000 | −97.7999 | 0.0000 | −97.7999 | 0.0000 | −97.7999 | 0.0000 |
| 2 | 64.1 | 64.1000 | 0.0000 | 64.1000 | 0.0000 | 64.1000 | 0.00001 | 64.1001 | 0.0001 |
| 3 | 483.1 | 483.1000 | 0.0000 | 483.1000 | 0.0000 | 483.1000 | 0.00002 | 502.5821 | 19.4821 |
| 4 | −96.7 | −96.6999 | 0.0000 | −96.6999 | 0.0000 | −96.6999 | 0.0000 | −96.6999 | 0.0000 |
| 5 | −395 | −394.9999 | 0.0000 | −394.9999 | 0.0000 | −394.9999 | 0.00004 | −394.9992 | 0.0007 |
| 6 | −34.6 | −34.5999 | 0.0000 | −34.5999 | 0.0000 | −34.5999 | 0.0000 | −34.5999 | 0.0000 |
| 7 | 494 | 494.0000 | 0.0000 | 494.0000 | 0.0000 | 494.0000 | 0.0000 | 494.0001 | 0.0001 |
| 8 | −402.2 | −402.1999 | 0.0000 | −402.1998 | 0.0001 | −402.1914 | 0.0085 | −402.0527 | 0.1472 |
| 9 | −97.8 | −97.7999 | 0.0000 | −97.7999 | 0.0000 | −97.7999 | 0.0000 | −97.7999 | 0.0000 |
| 10 | 64.1 | 64.1000 | 0.0000 | 64.1000 | 0.0000 | 64.1000 | 0.00001 | 64.1002 | 0.0002 |
| 11 | 483.1 | 483.1000 | 0.0000 | 483.1000 | 0.0000 | 483.1005 | 0.0005 | 507.0740 | 23.9740 |
| 12 | −96.7 | −96.6999 | 0.0000 | −96.6999 | 0.0000 | −96.6999 | 0.0000 | −96.6999 | 0.0000 |
| 13 | −395 | −394.9999 | 0.0000 | −394.9999 | 0.0000 | −394.9999 | 0.00006 | −394.9993 | 0.0006 |
| 14 | −34.6 | −34.5999 | 0.0000 | −34.5999 | 0.0000 | −34.5999 | 0.0000 | −34.5999 | 0.0000 |
| 15 | 494 | 494.0000 | 0.0000 | 494.0000 | 0.0000 | 494.0000 | 0.0000 | 494.0001 | 0.0001 |
| 16 | −402.2 | −402.1999 | 0.0000 | −402.1998 | 0.0001 | −402.1931 | 0.0068 | −402.0526 | 0.1473 |

**A9:** Results of multimodal functions for dimensions 2, 5, 10, and 20 for problem instance 10.

| Problem ID | $f^*$ | Dimension =2 | | Dimension=5 | | Dimension=10 | | Dimension=20 | |
|---|---|---|---|---|---|---|---|---|---|
| | | $f(x)$ | $\epsilon_f$ | $f(x)$ | $\epsilon_f$ | $f(x)$ | $\epsilon_f$ | $f(x)$ | $\epsilon_f$ |
| 1 | −97.8 | −97.7999 | 0.0000 | −97.7999 | 0.0000 | −97.7999 | 0.0000 | −97.7999 | 0.0000 |
| 2 | 64.1 | 64.1000 | 0.0000 | 64.1000 | 0.0000 | 64.1000 | 0.00001 | 64.1002 | 0.0002 |
| 3 | 483.1 | 483.1000 | 0.0000 | 483.1000 | 0.0000 | 483.1000 | 0.00004 | 506.6890 | 23.5890 |
| 4 | −96.7 | −96.6999 | 0.0000 | −96.6999 | 0.0000 | −96.6999 | 0.0000 | −96.6999 | 0.0000 |
| 5 | −395 | −394.9999 | 0.0000 | −394.9999 | 0.0000 | −394.9999 | 0.00009 | −394.9992 | 0.0007 |
| 6 | −34.6 | −34.5999 | 0.0000 | −34.5999 | 0.0000 | −34.5999 | 0.0000 | −34.5999 | 0.0000 |
| 7 | 494 | 494.0000 | 0.0000 | 494.0000 | 0.0000 | 494.0000 | 0.0000 | 494.0001 | 0.0001 |
| 8 | −402.2 | −402.1999 | 0.0000 | −402.1998 | 0.0001 | −402.1930 | 0.0069 | −402.0551 | 0.1448 |
| 9 | −97.8 | −97.7999 | 0.0000 | −97.7999 | 0.0000 | −97.7999 | 0.0000 | −97.7999 | 0.0000 |
| 10 | 64.1 | 64.1000 | 0.0000 | 64.1000 | 0.0000 | 64.1000 | 0.00001 | 64.1002 | 0.0002 |
| 11 | 483.1 | 483.1000 | 0.0000 | 483.1000 | 0.0000 | 483.1005 | 0.0005 | 491.1233 | 8.0233 |
| 12 | −96.7 | −96.6999 | 0.0000 | −96.6999 | 0.0000 | −96.6999 | 0.0000 | −96.6999 | 0.0000 |
| 13 | −395 | −394.9999 | 0.0000 | −394.9999 | 0.0000 | −394.9999 | 0.00006 | −394.9992 | 0.0007 |
| 14 | −34.6 | −34.5999 | 0.0000 | −34.5999 | 0.0000 | −34.5999 | 0.0000 | −34.5999 | 0.0000 |
| 15 | 494 | 494.0000 | 0.0000 | 494.0000 | 0.0000 | 494.0000 | 0.0000 | 494.0001 | 0.0001 |
| 16 | −402.2 | −402.1999 | 0.0000 | −402.1998 | 0.0001 | −402.1915 | 0.0084 | −402.0773 | 0.1226 |

**A10:** Results of multimodal functions for dimensions 2, 5, 10, and 20 for problem instance 11.

| Problem ID | $f^*$ | Dimension =2 | | Dimension=5 | | Dimension=10 | | Dimension=20 | |
|---|---|---|---|---|---|---|---|---|---|
| | | $f(x)$ | $\epsilon_f$ | $f(x)$ | $\epsilon_f$ | $f(x)$ | $\epsilon_f$ | $f(x)$ | $\epsilon_f$ |
| 1 | −97.8 | −97.7999 | 0.0000 | −97.7999 | 0.0000 | −97.7999 | 0.0000 | −97.7999 | 0.0000 |
| 2 | 64.1 | 64.1000 | 0.0000 | 64.1000 | 0.0000 | 64.1000 | 0.00001 | 64.1002 | 0.0002 |
| 3 | 483.1 | 483.1000 | 0.0000 | 483.1000 | 0.0000 | 483.1000 | 0.00002 | 503.1446 | 20.0446 |
| 4 | −96.7 | −96.6999 | 0.0000 | −96.6999 | 0.0000 | −96.6999 | 0.0000 | −96.6999 | 0.0000 |
| 5 | −395 | −394.9999 | 0.0000 | −394.9999 | 0.0000 | −394.9999 | 0.00007 | −394.9994 | 0.0005 |
| 6 | −34.6 | −34.5999 | 0.0000 | −34.5999 | 0.0000 | −34.5999 | 0.0000 | −34.5999 | 0.0000 |
| 7 | 494 | 494.0000 | 0.0000 | 494.0000 | 0.0000 | 494.0000 | 0.0000 | 494.0001 | 0.0001 |
| 8 | −402.2 | −402.1999 | 0.0000 | −402.1999 | 0.0000 | −402.1929 | 0.0070 | −402.0744 | 0.1255 |
| 9 | −97.8 | −97.7999 | 0.0000 | −97.7999 | 0.0000 | −97.7999 | 0.0000 | −97.7999 | 0.0000 |
| 10 | 64.1 | 64.1000 | 0.0000 | 64.1000 | 0.0000 | 64.1000 | 0.00001 | 64.1002 | 0.0002 |
| 11 | 483.1 | 483.1000 | 0.0000 | 483.1000 | 0.0000 | 483.1022 | 0.0022 | 495.4908 | 12.3908 |
| 12 | −96.7 | −96.6999 | 0.0000 | −96.6999 | 0.0000 | −96.6999 | 0.0000 | −96.6999 | 0.0000 |
| 13 | −395 | −394.9999 | 0.0000 | −394.9999 | 0.0000 | −394.9999 | 0.00005 | −394.9993 | 0.0006 |
| 14 | −34.6 | −34.5999 | 0.0000 | −34.5999 | 0.0000 | −34.5999 | 0.0000 | −34.5999 | 0.0000 |
| 15 | 494 | 494.0000 | 0.0000 | 494.0000 | 0.0000 | 494.0000 | 0.0000 | 494.0001 | 0.0001 |
| 16 | −402.2 | −402.1999 | 0.0000 | −402.1998 | 0.0002 | −402.1941 | 0.0058 | −402.0719 | 0.1280 |

**A11:** Results of multimodal functions for dimensions 2, 5, 10, and 20 for problem instance 12.

| Problem ID | $f^*$ | Dimension =2 | | Dimension=5 | | Dimension=10 | | Dimension=20 | |
|---|---|---|---|---|---|---|---|---|---|
| | | $f(x)$ | $\epsilon_f$ | $f(x)$ | $\epsilon_f$ | $f(x)$ | $\epsilon_f$ | $f(x)$ | $\epsilon_f$ |
| 1 | −97.8 | −97.7999 | 0.0000 | −97.7999 | 0.0000 | −97.7999 | 0.0000 | −97.7999 | 0.0000 |
| 2 | 64.1 | 64.1000 | 0.0000 | 64.1000 | 0.0000 | 64.1000 | 0.00001 | 64.1002 | 0.0002 |
| 3 | 483.1 | 483.1000 | 0.0000 | 483.1000 | 0.0000 | 483.1000 | 0.00004 | 509.3977 | 26.2977 |
| 4 | −96.7 | −96.6999 | 0.0000 | −96.6999 | 0.0000 | −96.6999 | 0.0000 | −96.6999 | 0.0000 |
| 5 | −395 | −394.9999 | 0.0000 | −394.9999 | 0.0000 | −394.9999 | 0.00006 | −394.9994 | 0.0005 |
| 6 | −34.6 | −34.5999 | 0.0000 | −34.5999 | 0.0000 | −34.5999 | 0.0000 | −34.5999 | 0.0000 |
| 7 | 494 | 494.0000 | 0.0000 | 494.0000 | 0.0000 | 494.0000 | 0.0000 | 494.0001 | 0.0001 |
| 8 | −402.2 | −402.1999 | 0.0000 | −402.1998 | 0.0001 | −402.1920 | 0.00798 | −402.0593 | 0.1406 |
| 9 | −97.8 | −97.7999 | 0.0000 | −97.7999 | 0.0000 | −97.7999 | 0.0000 | −97.7999 | 0.0000 |
| 10 | 64.1 | 64.1000 | 0.0000 | 64.1000 | 0.0000 | 64.1000 | 0.00001 | 64.1002 | 0.0002 |
| 11 | 483.1 | 483.1000 | 0.0000 | 483.1000 | 0.0000 | 483.1001 | 0.0001 | 490.1220 | 7.0220 |
| 12 | −96.7 | −96.6999 | 0.0000 | −96.6999 | 0.0000 | −96.6999 | 0.0000 | −96.6999 | 0.0000 |
| 13 | −395 | −394.9999 | 0.0000 | −394.9999 | 0.0000 | −394.9999 | 0.00006 | −394.9993 | 0.0006 |
| 14 | −34.6 | −34.5999 | 0.0000 | −34.5999 | 0.0000 | −34.5999 | 0.0000 | −34.5999 | 0.0000 |
| 15 | 494 | 494.0000 | 0.0000 | 494.0000 | 0.0000 | 494.0000 | 0.0000 | 494.0001 | 0.0001 |
| 16 | −402.2 | −402.1999 | 0.0000 | −402.1998 | 0.0001 | −402.1905 | 0.0094 | 402.0683 | 0.1316 |

**A12:** Results of multimodal functions for dimensions 2, 5, 10, and 20 for problem instance 13.

| Problem ID | $f^*$ | Dimension =2 | | Dimension=5 | | Dimension=10 | | Dimension=20 | |
|---|---|---|---|---|---|---|---|---|---|
| | | $f(x)$ | $\epsilon_f$ | $f(x)$ | $\epsilon_f$ | $f(x)$ | $\epsilon_f$ | $f(x)$ | $\epsilon_f$ |
| 1 | −97.8 | −97.7999 | 0.0000 | −97.7999 | 0.0000 | −97.7999 | 0.0000 | −97.7999 | 0.0000 |
| 2 | 64.1 | 64.1000 | 0.0000 | 64.1000 | 0.0000 | 64.1000 | 0.0000 | 64.1002 | 0.0002 |
| 3 | 483.1 | 483.1000 | 0.0000 | 483.1000 | 0.0000 | 483.1000 | 0.00004 | 519.0001 | 35.9001 |
| 4 | −96.7 | −96.6999 | 0.0000 | −96.6999 | 0.0000 | −96.6999 | 0.0000 | −96.6999 | 0.0000 |
| 5 | −395 | −394.9999 | 0.0000 | −394.9999 | 0.0000 | −394.9999 | 0.00006 | −394.9993 | 0.0006 |
| 6 | −34.6 | −34.5999 | 0.0000 | −34.5999 | 0.0000 | −34.5999 | 0.0000 | −34.5999 | 0.0000 |
| 7 | 494 | 494.0000 | 0.0000 | 494.0000 | 0.0000 | 494.0000 | 0.0000 | 494.0001 | 0.0001 |
| 8 | −402.2 | −402.1999 | 0.0000 | −402.1998 | 0.0001 | −402.0692 | 0.0099 | −402.0692 | 0.1307 |
| 9 | −97.8 | −97.7999 | 0.0000 | −97.7999 | 0.0000 | −97.7999 | 0.0000 | −97.7999 | 0.0000 |
| 10 | 64.1 | 64.1000 | 0.0000 | 64.1000 | 0.0000 | 64.1000 | 0.00002 | 64.1002 | 0.0002 |
| 11 | 483.1 | 483.1000 | 0.0000 | 483.1000 | 0.0000 | 483.1009 | 0.0009 | 508.8228 | 25.7228 |
| 12 | −96.7 | −96.6999 | 0.0000 | −96.6999 | 0.0000 | −96.6999 | 0.0000 | −96.6999 | 0.0000 |
| 13 | −395 | −394.9999 | 0.0000 | −394.9999 | 0.0000 | −394.9999 | 0.00007 | −394.9993 | 0.0006 |
| 14 | −34.6 | −34.5999 | 0.0000 | −34.5999 | 0.0000 | −34.5999 | 0.0000 | −34.5999 | 0.0000 |
| 15 | 494 | 494.0000 | 0.0000 | 494.0000 | 0.0000 | 494.0000 | 0.0000 | 494.0001 | 0.0001 |
| 16 | −402.2 | −402.1999 | 0.0000 | −402.1998 | 0.0001 | −402.1911 | 0.0088 | −402.0825 | 0.1174 |

**A13:** Results of multimodal functions for dimensions 2, 5, 10, and 20 for problem instance 14.

| Problem ID | $f^*$ | Dimension =2 | | Dimension=5 | | Dimension=10 | | Dimension=20 | |
|---|---|---|---|---|---|---|---|---|---|
| | | $f(x)$ | $\epsilon_f$ | $f(x)$ | $\epsilon_f$ | $f(x)$ | $\epsilon_f$ | $f(x)$ | $\epsilon_f$ |
| 1 | −97.8 | −97.7999 | 0.0000 | −97.7999 | 0.0000 | −97.7999 | 0.0000 | −97.7999 | 0.0000 |
| 2 | 64.1 | 64.1000 | 0.0000 | 64.1000 | 0.0000 | 64.1000 | 0.00001 | 64.1002 | 0.0002 |
| 3 | 483.1 | 483.1000 | 0.0000 | 483.1000 | 0.0000 | 483.1023 | 0.0023 | 502.7916 | 19.6916 |
| 4 | −96.7 | −96.6999 | 0.0000 | −96.6999 | 0.0000 | −96.6999 | 0.0000 | −96.6999 | 0.0000 |
| 5 | −395 | −394.9999 | 0.0000 | −394.9999 | 0.0000 | −394.9999 | 0.00008 | −394.9994 | 0.0005 |
| 6 | −34.6 | −34.5999 | 0.0000 | −34.5999 | 0.0000 | −34.5999 | 0.0000 | −34.5999 | 0.0000 |
| 7 | 494 | 494.0000 | 0.0000 | 494.0000 | 0.0000 | 494.0000 | 0.0000 | 494.0001 | 0.0001 |
| 8 | −402.2 | −402.1999 | 0.0000 | −402.1999 | 0.0000 | −402.1929 | 0.0070 | −402.0631 | 0.1368 |
| 9 | −97.8 | −97.7999 | 0.0000 | −97.7999 | 0.0000 | −97.7999 | 0.0000 | −97.7999 | 0.0000 |
| 10 | 64.1 | 64.1000 | 0.0000 | 64.1000 | 0.0000 | 64.1000 | 0.00002 | 64.1002 | 0.0002 |
| 11 | 483.1 | 483.1000 | 0.0000 | 483.1000 | 0.0000 | 483.1000 | 0.0000 | 484.9494 | 1.8494 |
| 12 | −96.7 | −96.6999 | 0.0000 | −96.6999 | 0.0000 | −96.6999 | 0.0000 | −96.6999 | 0.0000 |
| 13 | −395 | −394.9999 | 0.0000 | −394.9999 | 0.0000 | −394.9999 | 0.00007 | −394.9994 | 0.0005 |
| 14 | −34.6 | −34.5999 | 0.0000 | −34.5999 | 0.0000 | −34.5999 | 0.0000 | −34.5999 | 0.0000 |
| 15 | 494 | 494.0000 | 0.0000 | 494.0000 | 0.0000 | 494.0000 | 0.0000 | 494.0001 | 0.0001 |
| 16 | −402.2 | −402.1999 | 0.0000 | −402.1998 | 0.0001 | −402.1927 | 0.0072 | −402.0919 | 0.1080 |

**A14:** Results of multimodal functions for dimensions 2, 5, 10, and 20 for problem instance 15.

| Problem ID | $f^*$ | Dimension =2 | | Dimension=5 | | Dimension=10 | | Dimension=20 | |
|---|---|---|---|---|---|---|---|---|---|
| | | $f(x)$ | $\epsilon_f$ | $f(x)$ | $\epsilon_f$ | $f(x)$ | $\epsilon_f$ | $f(x)$ | $\epsilon_f$ |
| 1 | −97.8 | −97.7999 | 0.0000 | −97.7999 | 0.0000 | −97.7999 | 0.0000 | −97.7999 | 0.0000 |
| 2 | 64.1 | 64.1000 | 0.0000 | 64.1000 | 0.0000 | 64.1000 | 0.00001 | 64.1001 | 0.0001 |
| 3 | 483.1 | 483.1000 | 0.0000 | 483.1000 | 0.0000 | 483.1004 | 0.0004 | 483.1427 | 0.0427 |
| 4 | −96.7 | −96.6999 | 0.0000 | −96.6999 | 0.0000 | −96.6999 | 0.0000 | −96.6999 | 0.0000 |
| 5 | −395 | −394.9999 | 0.0000 | −394.9999 | 0.0000 | −394.9999 | 0.00006 | −394.9993 | 0.0006 |
| 6 | −34.6 | −34.5999 | 0.0000 | −34.5999 | 0.0000 | −34.5999 | 0.0000 | −34.5999 | 0.0000 |
| 7 | 494 | 494.0000 | 0.0000 | 494.0000 | 0.0000 | 494.0000 | 0.0000 | 494.0001 | 0.0001 |
| 8 | −402.2 | −402.1999 | 0.0000 | −402.1998 | 0.0001 | −402.1942 | 0.0057 | −402.0629 | 0.1370 |
| 9 | −97.8 | −97.7999 | 0.0000 | −97.7999 | 0.0000 | −97.7999 | 0.0000 | −97.7999 | 0.0000 |
| 10 | 64.1 | 64.1000 | 0.0000 | 64.1000 | 0.0000 | 64.1000 | 0.00002 | 64.1000 | 0.00008 |
| 11 | 483.1 | 483.1000 | 0.0000 | 483.1000 | 0.0000 | 483.1000 | 0.0000 | 505.6261 | 22.5261 |
| 12 | −96.7 | −96.6999 | 0.0000 | −96.6999 | 0.0000 | −96.6999 | 0.0000 | −96.6999 | 0.0000 |
| 13 | −395 | −394.9999 | 0.0000 | −394.9999 | 0.0000 | −394.9999 | 0.00005 | −394.9994 | 0.0005 |
| 14 | −34.6 | −34.5999 | 0.0000 | −34.5999 | 0.0000 | −34.5999 | 0.0000 | −34.5999 | 0.0000 |
| 15 | 494 | 494.0000 | 0.0000 | 494.0000 | 0.0000 | 494.0000 | 0.0000 | 494.0001 | 0.0001 |
| 16 | −402.2 | −402.1999 | 0.0000 | −402.1999 | 0.0001 | 402.1909 | 0.0090 | −402.0646 | 0.1353 |